%% file: neurips_2026.tex
\documentclass{article}

\usepackage[preprint,main]{neurips_2026}

\usepackage[utf8]{inputenc}
\usepackage[T1]{fontenc}

\usepackage[hidelinks]{hyperref}
\usepackage{url}
\usepackage{booktabs}
\usepackage{tabularx}
\usepackage{enumitem}
\usepackage{amsfonts}
\usepackage{amsmath}
\usepackage{amssymb}
\usepackage{nicefrac}
\usepackage{microtype}
\usepackage[table]{xcolor}
\usepackage{graphicx}
\usepackage{wrapfig}

\title{One Scan is Enough: Demonstration-Free Adaptation for Language-Guided Navigation}

\author{%
  Zhuoyuan Yu\thanks{Equal Contribution.}\quad
  Yuxing Long\footnotemark[1]\quad
  Du Wang\footnotemark[1]\\
  \textbf{Junyang Wang}\quad
  \textbf{Hongwei Fan}\quad
  \textbf{Hao Dong}\thanks{Corresponding author: \texttt{hao.dong@pku.edu.cn}.}\\
  \normalfont CFCS, School of Computer Science, Peking University\\
  \normalfont PrimeBot
}

\begin{document}

\maketitle
\suppressfloats[t]

\input{sections/1_abstract}

\input{sections/2_introduction}

\input{sections/3_related_work}

\input{sections/4_method}

\input{sections/5_experiments}

\input{sections/6_limitations}

\clearpage

\begin{ack}
This work was supported by the National Natural Science Foundation of China (Grant No. 62136001) and the Beijing Natural Science Foundation (Grant No. L2608141).
\end{ack}

\bibliographystyle{plainnat}
\bibliography{neurips_2026_refs}

\end{document}

%% file: sections/1_abstract.tex
\begin{abstract}
Deploying a language-guided navigation policy in a new building usually means watching it fail. Public-benchmark VLAs degrade scene by scene as layout, appearance, and language drift, and the standard fix of teleoperating fresh trajectories on site does not scale. We replace human demonstration with a single brief iPhone scan. \textbf{Scan2Instruction} compiles that scan into a complete training environment, automatically synthesizing executable navigation episodes with grounded instructions. \textbf{PRPT-KS} then adapts the policy through a closed-form Pareto utility read directly off the reconstructed scene, with no learned critic and no human labels. Across five real environments, the two designs together drive our RealNav to a \textbf{85\% real-robot success rate}, compared with \textbf{49\%} for a RealNav baseline post-trained on human demonstrations, and still achieve \textbf{82\% in unscanned regions of the same buildings}, compared with \textbf{41\%} for the same baseline. Our work turns demonstration-free scene adaptation for language-guided navigation from an aspiration into a working reality.
\end{abstract}

%% file: sections/2_introduction.tex
\section{Introduction}

Mobile robots become most useful when users can simply talk to them, with requests like \emph{go to the printer along the corridor}, \emph{find the white mug in the kitchen}, or \emph{look for somewhere to sit}. Modern vision-and-language navigation policies perform reasonably on benchmark trajectories, but performance falls off sharply once they leave those benchmarks: every new building changes both what the robot sees and what users tend to ask. Adapting to each target space by collecting real-robot demonstrations is slow, expensive, and does not scale to the long tail of homes, offices, and institutions. This paper asks: \textbf{\emph{Can a robot adapt to a new scene from a single passive scan, without any demonstrations?}}

\begin{figure}[t]
  \centering
  \includegraphics[width=1.0\linewidth]{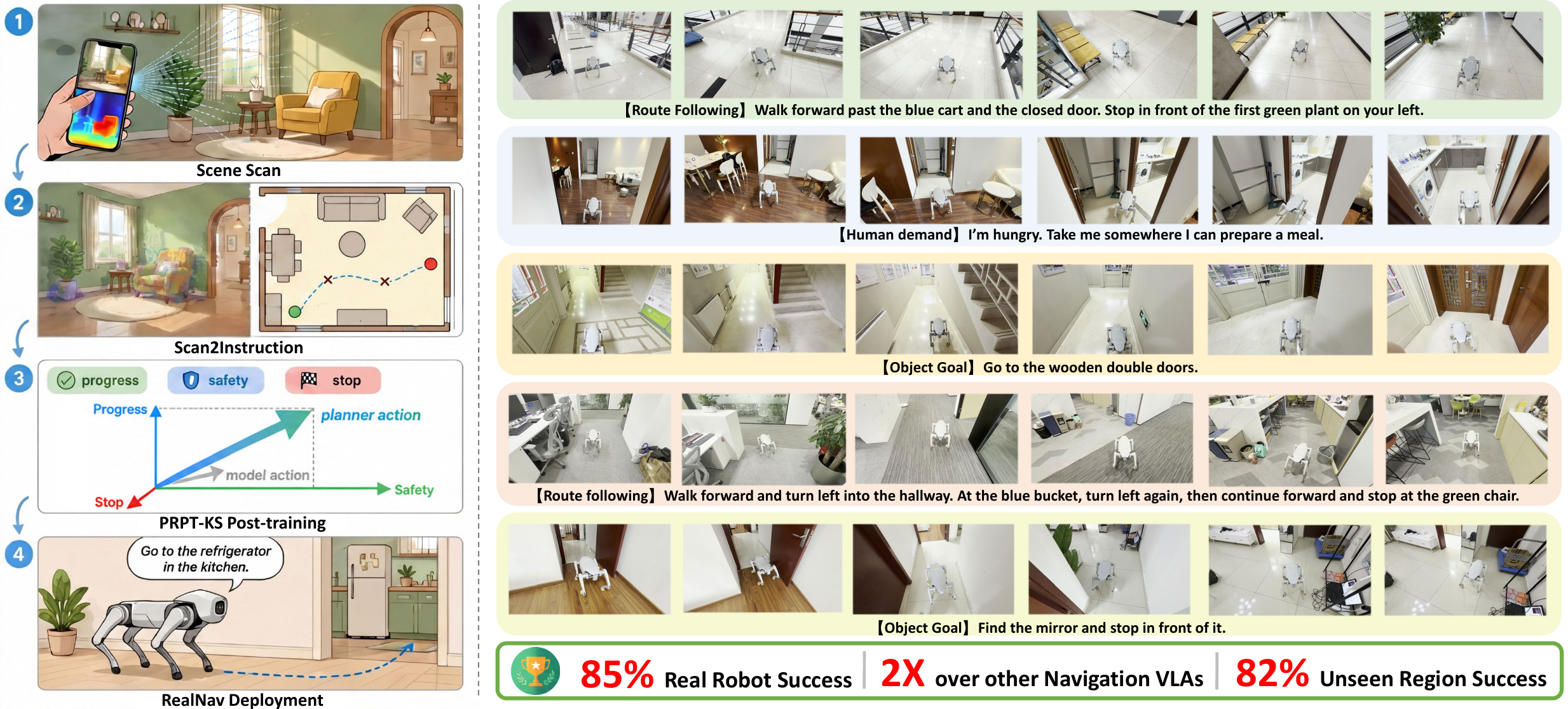}
  \caption{\textbf{One iPhone scan, a deployable language-guided navigation policy.} \textbf{Left:} \textbf{Scan2Instruction} compiles a brief handheld scan into scene-specific training episodes, and \textbf{PRPT-KS} adapts \textbf{RealNav} on them with no demonstrations and no human labels. \textbf{Right:} real-robot rollouts following natural-language instructions in the scanned scenes.}
  \label{fig:teaser}
\end{figure}

Recent advances in 3D Gaussian Splatting~\citep{kerbl2023gaussian} reconstruct real environments from a casual scan and re-render photorealistic first-person observations inside them. A line of real-to-sim-to-real navigation systems builds on this capability to adapt a navigation policy directly in the reconstructed target scene~\citep{embodedsplat2025,navgsim2026,vrrobo2025,vid2sim2025}. These systems demonstrate the value of reconstructed environments for navigation policy adaptation. Building on this foundation, we study how a single passive scan can automatically supply both grounded language-navigation episodes and geometry-derived post-training signals, without human-provided task demonstrations in the target scene. Figure~\ref{fig:teaser} shows the overall pipeline.

The first component, \textbf{Scan2Instruction}, reframes a scanned scene from a passive renderer into an active source of training tasks. From the scan we extract both a Gaussian rendering of the environment and a queryable geometric map of free space, and then sample feasible navigation episodes spanning three complementary instruction types: Route following, Object Goal, and Human Demand. For each episode, the compiler plans a collision-aware reference trajectory, renders first-person observations along it, and synthesizes natural-language instructions that are visually grounded in the rendered viewpoints and spatially aligned with the planned route. After a passive scan and one-time manual scene calibration, task generation and adaptation require no human demonstrations.

The second component, \textbf{PRPT-KS}, is a scene-grounded reinforcement post-training algorithm for our continuous-trajectory navigation VLA \textbf{RealNav}. Once the scene is known, standard RL's expensive components become unnecessary: the learned critic, a learned reward model, large-scale exploration, and human preference labels. PRPT-KS reads a three-dimensional Pareto utility vector of path progress, obstacle safety, and correct stopping directly off the known scene, and instantiates the same vector three ways: as a scalar projection that drives the policy gradient, as a per-axis admissibility test that lets a planner intervene on unsafe or low-progress actions, and as a vector dominance check that re-weights preference pairs recovered from those interventions. Adaptation thus reaches strong performance using only scene-grounded signals---no task demonstrations, learned critic, or human labels in the target environment.

We evaluate on five real scenes (office, cottage, teaching building, two apartments) in both Gaussian simulation and on the robot. The scene-adapted RealNav reaches a \textbf{85\% real-robot success rate on scanned regions, compared with 49\% for the same model post-trained on human demonstrations, and generalizes to never-scanned regions at 82\% versus 41\% for the same baseline}.

Our contributions are threefold:
\begin{itemize}[leftmargin=*,itemsep=2pt,topsep=2pt,parsep=0pt]
  \item We propose \textbf{Scan2Instruction}, the first training-data compiler that turns a single passive iPhone scan into scene-specific episodes with grounded language instructions, removing the per-scene teleoperation-and-annotation bottleneck of language-guided navigation.
  \item We design \textbf{PRPT-KS}, a scene-grounded RL post-training algorithm that reads a three-axis Pareto utility off the known scene and reuses it as policy gradient, planner admissibility test, and preference re-weighting, replacing the learned critic, learned reward model, exploration, and human labels.
  \item \textbf{Experiments} on five real scenes show that they drive \textbf{RealNav} to \textbf{85\%} real-robot success on scanned regions and \textbf{82\%} on unscanned ones, compared with 49\% / 41\% for the baseline post-trained on human demonstrations.
\end{itemize}

%% file: sections/3_related_work.tex
\section{Related Work}

\subsection{3D Gaussian Splatting for Embodied Navigation}

3D Gaussian Splatting (3DGS) reconstructs real scenes and renders robot-centric observations, reducing the sim-to-real visual gap for embodied navigation \citep{kerbl2023gaussian,embodedsplat2025,navgsim2026,vrrobo2025,vid2sim2025}. We share the reconstruct-and-render setting with this line of work. Prior systems use reconstructed scenes for simulation and navigation policy training. Our focus is to automatically compile a single passive scan into grounded language-action episodes and geometry-derived post-training signals. Scan2Instruction samples executable tasks, plans collision-aware trajectories, renders first-person observations, and generates diverse instructions grounded in those trajectories. Thus, the scene becomes both a simulator and a source of language-action episodes and geometry-derived post-training signals for our continuous-trajectory VLA RealNav.

\subsection{Reinforcement Post-Training for Navigation Policies}

Reinforcement learning is widely used for embodied navigation, with training signals typically obtained through massive simulator interaction~\citep{wijmans2020ddppo}, map-based priors~\citep{chaplot2020activeslam,chaplot2020semexp}, target-driven objectives~\citep{zhu2017targetdriven}, or cross-modal reward shaping for VLN~\citep{wang2019rcm,ma2019selfmonitoring}. A complementary line uses real-world experience and expert interventions for policy improvement~\citep{ross2011dagger,kahn2021badgr,kahn2021land}, often combined with general-purpose optimization and preference-learning tools~\citep{schulman2017ppo,ouyang2022instructgpt,rafailov2023dpo}. These methods explore from scratch, depend on discrete graphs with expert paths, or require real-robot trial-and-error. In contrast, since the target scene is already scanned, we query free space, reference paths, obstacle distances, and goal regions directly from geometry. PRPT-KS converts this geometry into a closed-form Pareto utility along three axes (progress, safety, stopping), used as a scalar advantage replacing the learned critic, an admissibility test gating planner intervention, and a dominance weight on the resulting preference pairs. Together they adapt RealNav without human takeover, reward models, or learned per-scene critics.

%% file: sections/4_method.tex
\section{Method}
\label{sec:method}

\subsection{Problem formulation}
\label{subsec:demofree}

\paragraph{Task Setting.}
A mobile robot is deployed in a target environment for which the only prior input is a single passive scan $\mathcal{S}$ of the scene. At test time the robot receives natural-language instructions $i \in \mathcal{I}$ from the user. Adaptation is \emph{demonstration-free}: the scan $\mathcal{S}$ may be used freely, but no human-provided task trajectory is available in the target scene.

\paragraph{Optimization Objective.}
Let $\tau = (o_{1:T}, a_{1:T})$ denote a rollout produced by a navigation policy $\pi_\theta(a_t \mid o_{1:t}, i)$ in the target environment, and let $R(\tau, i) \in \{0, 1\}$ indicate whether the trial succeeds for instruction $i$. We seek an \emph{adaptation algorithm} $\mathcal{A}: (\theta_0, \mathcal{S}) \mapsto \theta$ such that
\begin{equation}
  \pi_\theta^\star
  \;=\;
  \arg\max_{\theta}\;
  \mathbb{E}_{i \sim \mathcal{I}}\,
  \mathbb{E}_{\tau \sim \pi_\theta(\cdot \mid \mathcal{S},\, i)}\!
  \left[\,R(\tau,\, i)\,\right]
  \quad\text{s.t.}\quad
  \theta = \mathcal{A}\!\left(\theta_0,\, \mathcal{S}\right),
\end{equation}
where $\theta_0$ are the parameters of a base navigation VLA pretrained without target-scene access.

\subsection{RealNav: A Continuous-Trajectory Navigation VLA}
\label{subsec:realnav}

\paragraph{Model architecture.}
RealNav is a continuous-trajectory navigation VLA built on Qwen3-VL~\citep{qwen3vl2025}. At each decision step $t$, the input is a natural-language instruction $i$ together with a sampled history of RGB observations $o_{1:t}$, and we write $s_t = (o_{1:t}, i)$ for the full conditioning. The instruction and observation tokens are fed through the Qwen3-VL chat backbone, and after the user turn we append a small set of dedicated \emph{navigation tokens}. These navigation tokens absorb the cross-modal context through self-attention, and their final hidden states are read out by a lightweight MLP \emph{action head}. Concretely, RealNav implements the deterministic mapping
\begin{equation}
  \mu_\theta(s_t)
  \;=\;
  f_{\mathrm{head}}\!\Big(\,h_{\mathrm{nav}}\!\left(s_t;\, \theta_{\mathrm{vlm}}\right)\,\Big)
  \;\in\;\mathbb{R}^{3K},
\end{equation}
where $h_{\mathrm{nav}}(\cdot; \theta_{\mathrm{vlm}})$ collects the final hidden states of the appended navigation tokens after passing $s_t$ through the Qwen3-VL backbone, and $f_{\mathrm{head}}$ is the MLP action head. The 24-dimensional output is reshaped into $K=8$ future local waypoints $\{(x_k, y_k, \theta_k)\}_{k=1}^{K}$ in the current robot frame, with $(x_k, y_k)$ a planar translation and $\theta_k$ a heading change. Stop is inferred when consecutive executed waypoint differences fall below preset position and heading thresholds, and is correct only when the robot is within the target radius.

\paragraph{Navigation pretraining.}
To establish a navigation-capable initial policy before any target-scene adaptation, we train $\mu_\theta$ on R2R-CE and RxR-CE~\citep{anderson2018vln,ku2020rxr,beyondnavgraph2020} by converting their reference trajectories into local waypoint labels $a_t^\star$ and fitting an MSE behavior-cloning loss directly on the deterministic output.

\paragraph{Gaussian policy parameterization for post-training.}
Both the policy-gradient ratio and the preference log-likelihood introduced in Section~\ref{subsec:erptks} treat the waypoint vector as a Gaussian $\pi_\theta(a_t \mid s_t) = \mathcal{N}(\mu_\theta(s_t), \Sigma)$ around the deterministic head, with isotropic covariance $\Sigma$ in normalized waypoint coordinates.

\subsection{Scan2Instruction: Compiling a scan into language-navigation episodes}
\label{subsec:compiler}

\begin{figure}[t]
  \centering
  \includegraphics[width=1.0\linewidth]{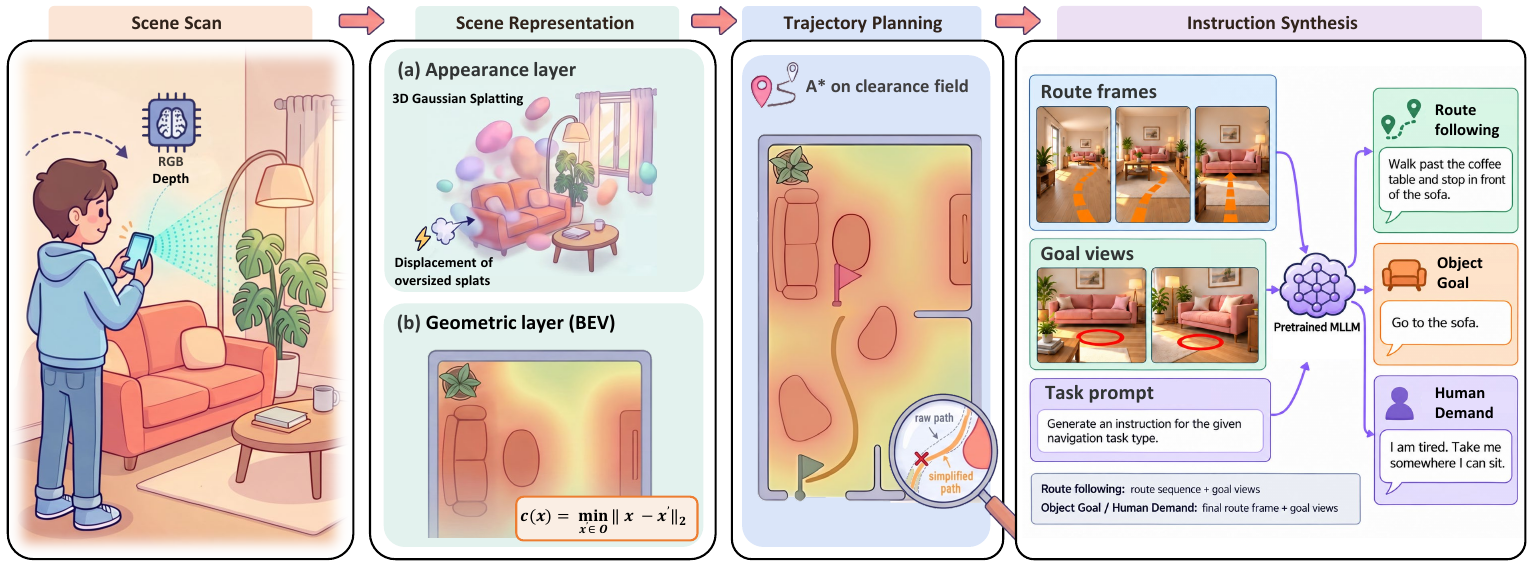}
  \caption{\textbf{Scan2Instruction pipeline.} A pretrained MLLM takes rendered images, goal views, and a type-specific prompt to generate Route following, Object Goal, or Human Demand instructions.}
  \label{fig:Scan2Instruction}
\end{figure}

Scan2Instruction turns a single scan into a dataset $\mathcal{D}_{\mathrm{scene}}$ of executable language-navigation episodes for the target environment. Figure~\ref{fig:Scan2Instruction} summarizes this stage.

\paragraph{Scene representation.}
The scan $\mathcal{S}$ is captured as an RGB-D video on a handheld iPhone with a built-in LiDAR sensor, with per-frame camera poses provided by ARKit or estimated offline using ORB-SLAM3~\citep{campos2021orbslam3}. From the posed RGB-D scan we recover a joint scene representation with two coupled layers. The \textbf{appearance layer} is a planar-regularized 3D Gaussian splat~\citep{kerbl2023gaussian,chen2024pgsr} reconstructed with LiDAR depth supervision. Gaussians with oversized projected footprints are moved far along the viewing ray to prevent them from occluding the scene and blurring the rendered view. The \textbf{geometric layer} is built by slicing an obstacle-height band from Gaussian occupancy projections, followed by thresholding, morphological cleanup, and connectivity filtering to obtain a reachable free-space mask. After one-time manual map alignment and traversability confirmation, we compute a continuous \emph{clearance field}
\begin{equation}
  c(\mathbf{x}) \;=\; \min_{\mathbf{x}' \in \mathcal{O}}\, \|\mathbf{x} - \mathbf{x}'\|_2,
  \label{eq:clearance}
\end{equation}
where $\mathcal{O}$ denotes the blocked cells of the eroded planning map $\mathcal{M}$ defined below, so that $c(\mathbf{x})$ is the distance to the boundary of $\mathcal{M}$. The clearance field $c(\cdot)$ supplies a continuous geometric quantity for trajectory planning below. The eroded free-space planning map $\mathcal{M}$, defined below, supplies the binary membership check inside $U_{\mathrm{safe}}$ (Eq.~\eqref{eq:u-safe}) and in the admissibility test of Section~\ref{subsec:erptks}.

\paragraph{Trajectory planning.}
Each episode begins with an executable reference trajectory planned purely from geometry, providing a concrete path for the language synthesis stage to describe.

We sample starts from $\mathcal{M}$ and goals from the same connected component of $\mathcal{M}$; a goal with $c(\mathbf{x})>c_{\mathrm{goal}}$ is moved to the nearest cell in that component with $c(\mathbf{x})\le c_{\mathrm{goal}}$, so that it lies next to an object that an instruction can name. Between each retained pair we run $A^\star$ with a clearance-penalized edge cost,
\begin{equation}
  \mathrm{cost}(\mathbf{x} \to \mathbf{x}')
  \;=\;
  s(\mathbf{x},\mathbf{x}') + \frac{w}{\max(d,\varepsilon)}\,m(d),
  \quad m(d)=\begin{cases}d_{\min}/(d+\varepsilon),&d<d_{\min},\\1,&\text{otherwise},\end{cases}
  \label{eq:astar-cost}
\end{equation}
where $s$ is the grid-step length, $d$ is obstacle distance in grid units, $w$ weights the clearance penalty, $d_{\min}$ sets the preferred clearance, and $\varepsilon>0$ stabilizes the denominator. The reachable free-space mask is eroded to form the planning map $\mathcal{M}$ with a safety margin; endpoints and traversed cells must remain inside $\mathcal{M}$ and satisfy clearance constraints. Safety-constrained RDP~\citep{douglas1973rdp} retains only chords that maintain minimum clearance. Accepted pairs are sampled in both directions.

\paragraph{Instruction synthesis.}
Scan2Instruction uses a pretrained multimodal large language model (MLLM) to synthesize three instruction types:
\begin{itemize}[leftmargin=*,itemsep=1pt,topsep=2pt,parsep=0pt]
  \item \textbf{Route following} --- actions and landmarks describing the route;
  \item \textbf{Object Goal} --- a named target object, without route directions;
  \item \textbf{Human Demand} --- a functional request, without naming the target object or describing the route.
\end{itemize}
The MLLM performs visual grounding and language generation: it interprets the rendered scene, identifies the destination, and expresses the task according to a type-specific text prompt. Route frames carry an orange future-path trail, while goal views mark the destination with a red ground circle. For instruction type $\gamma$, generation uses a single MLLM call:
\begin{equation}
 i=\mathrm{MLLM}\!\left(\{\tilde{o}_t\}_{t\in\mathcal{T}_\gamma},\,\mathcal{G};\,P_\gamma\right),
 \label{eq:instruction-generation}
\end{equation}
where $\tilde{o}_t$ are guided route images, $\mathcal{G}$ contains the goal views, and $P_\gamma$ specifies the instruction type and wording constraints. Route following uses the route sequence and goal views; Object Goal and Human Demand use goal views and the last route frame. 

For each planned route, we pair the MLLM-generated instruction with the planner's reference path, poses, and local waypoint labels. We render a separate sequence of policy observations with camera intrinsics and heights different from those used for instruction synthesis, omitting the visual guides. Each compiled episode therefore combines the language instruction, policy observations, and planner supervision as
\begin{equation}
  e \,=\, \big(\, i,\, \{o_t\}_{t=1}^{T},\, \{p_t\}_{t=1}^{T},\, \{a_t^\star\}_{t=1}^{T},\, \tau^\star,\, \gamma \,\big),
  \label{eq:episode-tuple}
\end{equation}
with $i$ the instruction, $o_t$ a rendered RGB observation, $p_t$ a reference pose, $a_t^\star$ the local waypoint label, $\tau^\star$ the planner's reference path, and $\gamma$ the family identifier. The union $\mathcal{D}_{\mathrm{scene}}=\{e\}$ is the dataset Scan2Instruction hands to PRPT-KS.

For evaluation episodes, the synthesized instructions are additionally checked manually.

\begin{figure}[t]
  \centering
  \includegraphics[width=1.0\linewidth]{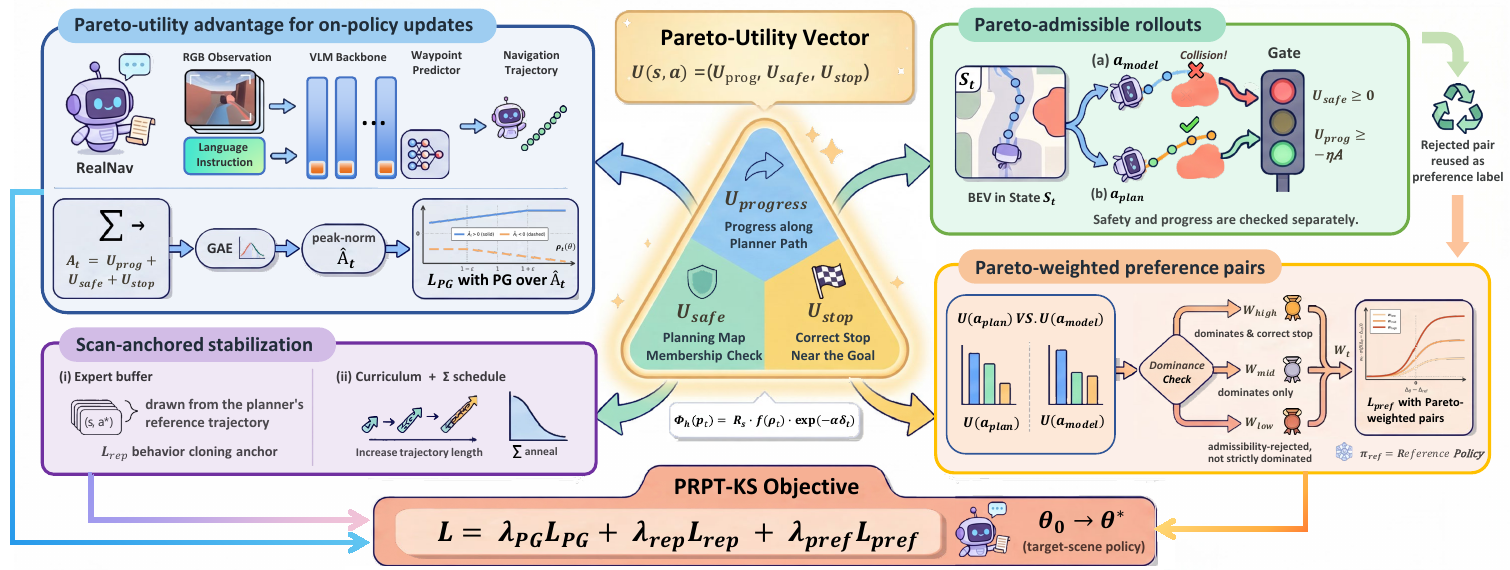}
  \caption{\textbf{PRPT-KS pipeline.} A Pareto utility vector read from known-scene geometry drives rollout updates, gates rollouts, and weights preference pairs.}
  \label{fig:PRPTKS}
\end{figure}

\subsection{PRPT-KS: Pareto-utility post-training for known-scene navigation}
\label{subsec:erptks}

\textbf{PRPT-KS}, short for \textbf{P}areto-utility \textbf{R}einforcement \textbf{P}ost-\textbf{T}raining in \textbf{K}nown \textbf{S}cenes, replaces the parts of standard RL that usually require expensive supervision with a single \emph{Pareto utility vector} read directly off the known scene. Concretely, every candidate action $a$ at state $s_t$ is scored by
\begin{equation}
  \mathbf{U}(s_t, a) \,=\, \big(\,U_{\mathrm{prog}}(s_t,a),\; U_{\mathrm{safe}}(s_t,a),\; U_{\mathrm{stop}}(s_t,a)\,\big),
  \label{eq:pareto-utility}
\end{equation}
whose three components quantify path progress, obstacle safety, and correct stopping respectively. Writing $\succ_P$ for the standard Pareto dominance on $\mathbb{R}^3$ (weak inequality on every axis and strict inequality on at least one), the remainder of this section instantiates PRPT-KS as three different read-outs of the \emph{same} $\mathbf{U}$: a scalar projection that drives the policy gradient (Eq.~\eqref{eq:advantage}), a per-axis admissibility test that gates planner intervention (Eq.~\eqref{eq:admissibility}), and a vector dominance check that re-weights pairwise preferences (Eq.~\eqref{eq:pareto-weight}). Because $\mathbf{U}$ is closed-form in known-scene geometry, none of these read-outs require a learned critic or reward model. Figure~\ref{fig:PRPTKS} illustrates the resulting training loop.

\paragraph{Three Pareto utilities from known-scene geometry.}
Let $\tau^\star$ be the planner's reference path for the current episode and $p_t$ the robot pose. The nearest reference-pose index $j_t$ among $N$ reference poses gives completion $\rho_t=j_t/(N-1)$; $\delta_t$ is distance to that pose. We use
\begin{equation}
 \Phi_h(p_t)=R_s f(\rho_t)\exp(-\alpha\bar\delta_t),\quad
 \bar\delta_t=\begin{cases}0,&\delta_t<\delta_{\mathrm{dead}},\\\delta_t,&\text{otherwise}.\end{cases}
 \label{eq:potential}
\end{equation}
Here $R_s>0$ scales the potential, $f$ is an increasing affine function of completion, $\alpha>0$ controls deviation decay, and $\delta_{\mathrm{dead}}$ defines a deviation dead zone. The utility components are
\begin{align}
 U_{\mathrm{prog}}(s_t,a)&=\gamma\Phi_h(p_{t+1}^{a})-\Phi_h(p_t),\label{eq:u-prog}\\
 U_{\mathrm{safe}}(s_t,a)&=-\lambda_{\mathrm{obs}}\,\mathbf{1}[\exists w\in\mathcal{E}(a):w\notin\mathcal{M}],\label{eq:u-safe}\\
 U_{\mathrm{stop}}(s_t,a)&=R_{\mathrm{succ}}\mathbf{1}[a\text{ is a stop and }\|p_{t+1}^{a}-g\|_2<r_{\mathrm{succ}}].\label{eq:u-stop}
\end{align}
Here $g$ is the goal position, and $\mathcal{E}(a)$ contains the interleaved execution subset of predicted waypoints, truncated at stopping. The safety penalty $\lambda_{\mathrm{obs}}>0$ applies when an executed waypoint leaves the eroded planning map $\mathcal{M}$, inside which every reference trajectory is planned. Training uses success reward $R_{\mathrm{succ}}>0$ and success radius $r_{\mathrm{succ}}$.

\paragraph{Pareto-utility advantage for rollout updates.}
The single-step training signal is a one-dimensional projection of $\mathbf{U}$ obtained by summing its components,
\begin{equation}
  A_t \,=\, U_{\mathrm{prog}}(s_t,a_t) + U_{\mathrm{safe}}(s_t,a_t) + U_{\mathrm{stop}}(s_t,a_t),
  \label{eq:advantage}
\end{equation}
We aggregate per-step signals with generalized advantage estimation~\citep{schulman2015gae} and peak-normalize the resulting advantages $\hat A_t$. Rollouts execute deterministic mean actions with planner substitution; all executed actions, including substitutions, enter a clipped importance-ratio update~\citep{schulman2017ppo}
\begin{equation}
  \mathcal{L}_{\mathrm{PG}}
  \,=\,
  -\,\mathbb{E}_t\!\Big[\,\min\!\big(\rho_t(\theta)\,\hat A_t,\ \mathrm{clip}\big(\rho_t(\theta),\,1{-}\epsilon,\,1{+}\epsilon\big)\,\hat A_t\big)\Big],
  \quad
  \rho_t(\theta) = \frac{\pi_\theta(a_t\mid s_t)}{\pi_{\mathrm{old}}(a_t\mid s_t)},
  \label{eq:pg-loss}
\end{equation}

\paragraph{Pareto-admissible rollouts.}
We wrap every step in a \emph{per-axis admissibility test}:
\begin{equation}
 \text{accept }a_t^{\mathrm{model}}\quad\Longleftrightarrow\quad
 U_{\mathrm{safe}}(s_t,a_t^{\mathrm{model}})\ge0
 \;\wedge\;
 U_{\mathrm{prog}}(s_t,a_t^{\mathrm{model}})\ge-\eta_A.
 \label{eq:admissibility}
\end{equation}
Here $\eta_A>0$ is the tolerance for progress regression. The test requires the checked execution waypoints to remain inside the eroded planning map $\mathcal{M}$ and prevents progress gains from compensating for a safety violation. Rejected actions are replaced by planner actions at the same state and retained as preference examples. Once the robot enters the training success radius, the planner controls the remaining steps to the endpoint and supplies further preference pairs.

\paragraph{Pareto-weighted preference pairs.}
Each rejected $a_t^{\mathrm{model}}$ is reused, paired with the planner's $a_t^{\mathrm{plan}}$ at the same state, as a state-conditioned preference label in continuous trajectory space. We consume these pairs with a pairwise preference objective~\citep{rafailov2023dpo} using the rollout-time policy as reference $\pi_{\mathrm{ref}}$. The per-pair weight $w_t$ is determined by \emph{vector} Pareto dominance of $a_t^{\mathrm{plan}}$ over $a_t^{\mathrm{model}}$:
\begin{equation}
  w_t = \begin{cases}
    w_{\mathrm{high}}, & \mathbf{U}(s_t,a_t^{\mathrm{plan}}) \succ_P \mathbf{U}(s_t,a_t^{\mathrm{model}}) \text{ and } U_{\mathrm{stop}}(s_t,a_t^{\mathrm{plan}}) > 0, \\
    w_{\mathrm{mid}},  & \mathbf{U}(s_t,a_t^{\mathrm{plan}}) \succ_P \mathbf{U}(s_t,a_t^{\mathrm{model}}), \\
    w_{\mathrm{low}},  & \text{otherwise (admissibility-rejected, not strictly dominated)},
  \end{cases}
  \label{eq:pareto-weight}
\end{equation}
with $w_{\mathrm{high}}>w_{\mathrm{mid}}>w_{\mathrm{low}}>0$. Combining $w_t$ with the standard sigmoid log-ratio formulation,
\begin{equation}
  \mathcal{L}_{\mathrm{pref}}
  \,=\,
  -\,\mathbb{E}\Big[\,w_t\,\log \sigma\!\big(\beta(\Delta_\theta-\Delta_{\mathrm{ref}})\big)\,\Big],
  \label{eq:dpo-loss}
\end{equation}
with $\beta$ the temperature and $\Delta_\theta$, $\Delta_{\mathrm{ref}}$ the policy and reference log-ratio gaps between $a_t^{\mathrm{plan}}$ and $a_t^{\mathrm{model}}$, computed under the same Gaussian parameterization of Section~\ref{subsec:realnav}.

\paragraph{Scan-anchored stabilization and unified objective.}
Three standard ingredients keep the post-training loop stable. A step-wise expert buffer of $(s, a^\star)$ transitions sampled along the planner's reference trajectories supplies a behavior-cloning anchor $\mathcal{L}_{\mathrm{rep}}$. A short-to-long trajectory-length curriculum decouples horizon ramp-up from policy improvement. A non-learned cosine schedule on $\Sigma$ controls action dispersion during post-training. The three Pareto read-outs jointly drive the unified post-training objective
\begin{equation}
  \mathcal{L}
  =
  \lambda_{\mathrm{PG}}\,\mathcal{L}_{\mathrm{PG}}
  +
  \lambda_{\mathrm{rep}}\,\mathcal{L}_{\mathrm{rep}}
  +
  \lambda_{\mathrm{pref}}\,\mathcal{L}_{\mathrm{pref}},
\end{equation}
where $\lambda_{\mathrm{PG}},\lambda_{\mathrm{rep}},\lambda_{\mathrm{pref}}>0$ balance the three terms, $\mathcal{L}_{\mathrm{PG}}$ is driven by the scalar projection of $\mathbf{U}$, $\mathcal{L}_{\mathrm{rep}}$ anchors the policy to planner reference actions, and $\mathcal{L}_{\mathrm{pref}}$ injects scene-grounded correction signals weighted by Pareto dominance.

%% file: sections/5_experiments.tex
\section{Experiments}
\label{sec:experiments}

\subsection{Setup}

\paragraph{Scenes and data.}
We evaluate on five real-world scenes: an office, a cottage, a teaching building, and two apartments. Each is scanned once and reconstructed into a renderable Gaussian scene and a geometric BEV map. Scan2Instruction then synthesizes the three instruction types of Section~\ref{subsec:compiler} per scan, with separately sampled training and evaluation episodes; disjoint routes or regions within a scanned scene are not guaranteed. Real-robot trial counts vary by scene, as summarized in Table~\ref{tab:scene_statistics}.

\input{tables/scene_statistics}

\paragraph{Baselines.}
We compare against one mapping baseline, \textbf{DualMap}~\cite{jiang2025dualmap}, which consumes the scan plus depth and pose streams and plans through a queryable open-vocabulary semantic map, and five recent VLN/VLA models: \textbf{NaVILA}~\cite{cheng2024navila}, \textbf{Uni-NaVid}~\cite{zhang2024uninavid}, \textbf{StreamVLN}~\cite{wei2025streamvln}, \textbf{CorrectNav}~\cite{yu2025correctnav}, and \textbf{OmniNav}~\cite{xue2025omninav}. All learning-based methods share the same RGB stream and low-level controller. We additionally report \textbf{RealNav}, our VLA pretrained on R2R-CE/RxR-CE, and demonstration-adapted variants \textbf{CorrectNav (human-demo-sft)} and \textbf{RealNav (human-demo-sft)}, fine-tuned on 20 teleoperated trajectories per scene.

\paragraph{Metrics.}
We report success rate (SR) in both simulation and on the physical robot, using a $1.5$\,m evaluation radius. SPL, navigation error (NE), and oracle success (OS) additionally require ground-truth path length and pose, which are not reliably measurable under our real-robot interface; we therefore report them only in the ablation.

\subsection{Simulation and real-world evaluation}
\input{tables/main_results}
Table~\ref{tab:main_results} reports success rate on the reconstructed Gaussian scenes and the physical robot under the same instruction types and task split. Representative real-robot rollouts appear in Figure~\ref{fig:teaser}~(right). RealNav is competitive with the external baselines in simulation but shows no clear advantage, indicating that generic navigation pre-training alone is insufficient for a newly scanned environment. After PRPT-KS post-training, success rate increases sharply across all five scenes despite the scene-specific experience coming from a single passive scan rather than demonstrations or teleoperation. The gain transfers to physical deployment: even with the rendering-camera gap and execution drift, the post-trained policy preserves strong performance, suggesting that PRPT-KS teaches the policy to follow language instructions while favoring safe, target-approaching actions under diverse viewpoints. The two human-demo-sft models, \textbf{CorrectNav (human-demo-sft)} and \textbf{RealNav (human-demo-sft)}, fine-tune their respective base models on the 20 teleoperated trajectories per scene; both improve over their untuned counterparts but remain far below RealNav~+~PRPT-KS.

\subsection{Generalization to unscanned regions of the same scenes}
\input{tables/unseen_area_results}
To test whether the policy memorizes the scanned region, we keep the same five environments but evaluate exclusively in regions that were \emph{never scanned}: not reconstructed into the Gaussian scene, never in the synthesized task set, and never visited by RL rollouts. We run 20 real-robot trials per scene (100 in total) under the same instruction types. Table~\ref{tab:unseen_area_results} reports real-robot SR in this setting. RealNav + PRPT-KS retains strong performance in these regions. These results suggest that the adapted policy generalizes beyond the scanned regions.

\subsection{Ablations}
\input{tables/ablation}
Table~\ref{tab:ablation} compares the full RealNav~+~PRPT-KS pipeline with the base RealNav policy before scene-specific post-training and with three loss-term ablations. \emph{w/o behavior-cloning anchor} sets $\lambda_{\mathrm{rep}}=0$, removing the planner-reference replay loss while retaining policy-gradient and preference updates. \emph{w/o Pareto-utility advantage} sets $\lambda_{\mathrm{PG}}=0$, disabling the policy-gradient term while retaining reference replay, preference supervision, and Pareto-admissible rollouts. \emph{w/o Pareto-weighted preference} sets $\lambda_{\mathrm{pref}}=0$, removing the entire weighted preference loss while retaining policy-gradient updates, reference replay, and rollout correction.

Scene-specific post-training increases SR from 0.23 to 0.85. Removing the behavior-cloning anchor causes the largest degradation, reducing SR to 0.30 and SPL from 0.834 to 0.293. Removing Pareto-utility advantage reduces SR to 0.81, SPL to 0.785, and OS from 0.88 to 0.85; removing Pareto-weighted preference reduces these metrics to 0.80, 0.781, and 0.83, respectively. These results support the contribution of all three loss terms to navigation success and path efficiency under the evaluated configuration. However, the variants without the policy-gradient term or preference supervision achieve lower NE (1.31\,m and 1.32\,m) than the full model (1.42\,m), so the improvements do not extend uniformly to terminal navigation error.

%% file: tables/scene_statistics.tex
\begin{table}[t]
  \centering
  \caption{\textbf{Target real-world scenes and synthesized data.} Evaluation instructions are manually checked.}
  \label{tab:scene_statistics}
  \footnotesize
  \setlength{\tabcolsep}{3pt}
  \renewcommand{\arraystretch}{1.05}
  \scalebox{1.0}{%
  \begin{tabular*}{\linewidth}{@{\extracolsep{\fill}}lccccc@{}}
    \toprule
    Scene & Area (m$^2$) & Scan (min) & Train eps. & Sim test & Real trials \\
    \midrule
    Office & 318 & 20 & 300 & 100 & 60 \\
    Cottage & 155 & 13 & 300 & 100 & 40 \\
    Teaching & 130 & 11 & 300 & 100 & 40 \\
    Apartment~1 & 90 & 9 & 300 & 100 & 30 \\
    Apartment~2 & 75 & 9 & 300 & 100 & 30 \\
    \midrule
    Total & 768 & 62 & 1500 & 500 & 200 \\
    \bottomrule
  \end{tabular*}%
  }
  \vspace{-0.2em}
\end{table}

%% file: tables/main_results.tex
\begin{table*}[t]
  \centering
  \caption{\textbf{Main evaluation on five scanned scenes.} Each cell reports success rate (SR) as \emph{sim / real} on reconstructed Gaussian scenes and the physical robot under the same task split. Average SR is computed from pooled successes and trials across scenes.}
  \label{tab:main_results}
  \footnotesize
  \setlength{\tabcolsep}{1pt}
  \renewcommand{\arraystretch}{1.05}
  \scalebox{1.0}{%
  \begin{tabularx}{\textwidth}{l>{\hsize=.92\hsize\linewidth=\hsize\centering\arraybackslash}X>{\hsize=.92\hsize\linewidth=\hsize\centering\arraybackslash}X>{\hsize=.92\hsize\linewidth=\hsize\centering\arraybackslash}X>{\hsize=1.16\hsize\linewidth=\hsize\centering\arraybackslash}X>{\hsize=1.16\hsize\linewidth=\hsize\centering\arraybackslash}X>{\hsize=.92\hsize\linewidth=\hsize\centering\arraybackslash}X}
    \toprule
    \textbf{Method} & \textbf{Office} & \textbf{Cottage} & \textbf{Teaching} & \mbox{\textbf{Apartment~1}} & \mbox{\textbf{Apartment~2}} & \textbf{Average} \\
    \midrule
    \multicolumn{7}{@{}l}{\textit{3D scene-graph mapping:}} \\
    \quad DualMap~\cite{jiang2025dualmap} & 0.32 / 0.17 & 0.22 / 0.13 & 0.17 / 0.10 & 0.35 / 0.17 & 0.33 / 0.17 & 0.28 / 0.15 \\
    \midrule
    \multicolumn{7}{@{}l}{\textit{Navigation VLA:}} \\
    \quad NaVILA~\cite{cheng2024navila} & 0.25 / 0.20 & 0.20 / 0.23 & 0.12 / 0.15 & 0.19 / 0.20 & 0.22 / 0.23 & 0.20 / 0.20 \\
    \quad Uni-NaVid~\cite{zhang2024uninavid} & 0.17 / 0.22 & 0.22 / 0.23 & 0.29 / 0.15 & 0.19 / 0.20 & 0.21 / 0.20 & 0.22 / 0.20 \\
    \quad StreamVLN~\cite{wei2025streamvln} & 0.26 / 0.17 & 0.26 / 0.13 & 0.10 / 0.13 & 0.23 / 0.30 & 0.30 / 0.23 & 0.23 / 0.18 \\
    \quad OmniNav~\cite{xue2025omninav} & 0.26 / 0.22 & 0.19 / 0.20 & 0.19 / 0.23 & 0.27 / 0.27 & 0.25 / 0.20 & 0.23 / 0.22 \\
    \quad CorrectNav~\cite{yu2025correctnav} & 0.13 / 0.18 & 0.14 / 0.23 & 0.18 / 0.25 & 0.36 / 0.27 & 0.29 / 0.20 & 0.22 / 0.22 \\
    \quad CorrectNav (human-demo-sft)~\cite{yu2025correctnav} & 0.35 / 0.47 & 0.36 / 0.43 & 0.40 / 0.48 & 0.58 / 0.57 & 0.51 / 0.50 & 0.44 / 0.48 \\
    \midrule
    \multicolumn{7}{@{}l}{\textit{Ours:}} \\
    \quad RealNav & 0.22 / 0.25 & 0.24 / 0.28 & 0.17 / 0.23 & 0.22 / 0.23 & 0.30 / 0.23 & 0.23 / 0.25 \\
    \quad RealNav (human-demo-sft) & 0.37 / 0.48 & 0.34 / 0.45 & 0.42 / 0.45 & 0.56 / 0.57 & 0.49 / 0.53 & 0.44 / 0.49 \\
    \rowcolor{black!10}
    \quad \textbf{RealNav + PRPT-KS} & \textbf{0.79 / 0.78} & \textbf{0.93 / 0.95} & \textbf{0.81 / 0.83} & \textbf{0.85 / 0.87} & \textbf{0.87 / 0.87} & \textbf{0.85 / 0.85} \\
    \bottomrule
  \end{tabularx}%
  }
  \vspace{-0.2em}
\end{table*}

%% file: tables/unseen_area_results.tex
\begin{table*}[t]
  \centering
  \caption{Real-robot evaluation in \emph{unscanned regions} of the same five scenes as Table~\ref{tab:main_results}: starts, goals, and intermediate waypoints all lie outside the scanned region used for reconstruction, task synthesis, and RL rollout. Each cell reports real-world success rate (SR).}
  \label{tab:unseen_area_results}
  \footnotesize
  \setlength{\tabcolsep}{1pt}
  \renewcommand{\arraystretch}{1.05}
  \scalebox{1.0}{%
  \begin{tabularx}{\textwidth}{l>{\hsize=.92\hsize\linewidth=\hsize\centering\arraybackslash}X>{\hsize=.92\hsize\linewidth=\hsize\centering\arraybackslash}X>{\hsize=.92\hsize\linewidth=\hsize\centering\arraybackslash}X>{\hsize=1.16\hsize\linewidth=\hsize\centering\arraybackslash}X>{\hsize=1.16\hsize\linewidth=\hsize\centering\arraybackslash}X>{\hsize=.92\hsize\linewidth=\hsize\centering\arraybackslash}X}
    \toprule
    \textbf{Method} & \textbf{Office} & \textbf{Cottage} & \textbf{Teaching} & \mbox{\textbf{Apartment~1}} & \mbox{\textbf{Apartment~2}} & \textbf{Average} \\
    \midrule
    \multicolumn{7}{@{}l}{\textit{3D scene-graph mapping:}} \\
    \quad DualMap~\cite{jiang2025dualmap} & 0.10 & 0.15 & 0.20 & 0.10 & 0.15 & 0.14 \\
    \midrule
    \multicolumn{7}{@{}l}{\textit{Navigation VLA:}} \\
    \quad NaVILA~\cite{cheng2024navila} & 0.15 & 0.20 & 0.20 & 0.10 & 0.25 & 0.18 \\
    \quad Uni-NaVid~\cite{zhang2024uninavid} & 0.20 & 0.30 & 0.15 & 0.20 & 0.25 & 0.22 \\
    \quad StreamVLN~\cite{wei2025streamvln} & 0.25 & 0.20 & 0.30 & 0.20 & 0.15 & 0.22 \\
    \quad OmniNav~\cite{xue2025omninav} & 0.30 & 0.25 & 0.20 & 0.35 & 0.20 & 0.26 \\
    \quad CorrectNav~\cite{yu2025correctnav} & 0.30 & 0.35 & 0.25 & 0.30 & 0.25 & 0.29 \\
    \quad CorrectNav (human-demo-sft)~\cite{yu2025correctnav} & 0.45 & 0.40 & 0.35 & 0.45 & 0.35 & 0.40 \\
    \midrule
    \multicolumn{7}{@{}l}{\textit{Ours:}} \\
    \quad RealNav & 0.25 & 0.30 & 0.20 & 0.20 & 0.15 & 0.22 \\
    \quad RealNav (human-demo-sft) & 0.45 & 0.40 & 0.35 & 0.45 & 0.40 & 0.41 \\
    \rowcolor{black!10}
    \quad \textbf{RealNav + PRPT-KS} & \textbf{0.80} & \textbf{0.85} & \textbf{0.80} & \textbf{0.85} & \textbf{0.80} & \textbf{0.82} \\
    \bottomrule
  \end{tabularx}%
  }
  \vspace{-0.2em}
\end{table*}

%% file: tables/ablation.tex
\begin{table*}[t]
  \centering
  \caption{\textbf{Ablation analysis in simulation.} We compare the full RealNav~+~PRPT-KS pipeline with RealNav before scene-specific post-training and with variants that individually disable one loss term in the unified post-training objective. Metrics are measured on reconstructed Gaussian scenes.}
  \label{tab:ablation}
  \footnotesize
  \setlength{\tabcolsep}{4pt}
  \renewcommand{\arraystretch}{1.05}
  
  \begin{tabularx}{\textwidth}{>{\raggedright\arraybackslash}p{0.52\textwidth}*{4}{>{\centering\arraybackslash}X}}
    \toprule
    \textbf{Variant} & \textbf{SR $\uparrow$} & \textbf{SPL $\uparrow$} & \textbf{NE (m) $\downarrow$} & \textbf{OS $\uparrow$} \\
    \midrule
    \rowcolor{black!10}
    \textbf{RealNav~+~PRPT-KS (Full)} & \textbf{0.85} & \textbf{0.834} & 1.42 & \textbf{0.88} \\
    \midrule
    RealNav (w/o PRPT-KS post-training) & 0.23 & 0.210 & 3.50 & 0.27 \\
    \midrule
    w/o behavior-cloning anchor & 0.30 & 0.293 & 3.49 & 0.41 \\
    w/o Pareto-utility advantage & 0.81 & 0.785 & \textbf{1.31} & 0.85 \\
    w/o Pareto-weighted preference & 0.80 & 0.781 & 1.32 & 0.83 \\
    \bottomrule
  \end{tabularx}
  
  \vspace{-0.2em}
\end{table*}

%% file: sections/6_limitations.tex
\section{Limitations and Broader Impacts}

\paragraph{Limitations and future work.}
Scanning still requires a human operator, and the geometry becomes stale when layouts change substantially, requiring a rescan. A natural extension is active, incremental mapping, where the robot detects divergence between reconstructed geometry and onboard observations and refreshes the representation autonomously.

\paragraph{Broader impacts.}
Replacing per-scene teleoperation with a single passive scan lowers the barrier to deploying language-guided robots in homes, offices, and small institutions. Two concerns warrant note: scans of private spaces may contain identifiable people or documents, so we release assets only after anonymization and require scanner consent from space owners; and stale geometry is a safety consideration in highly dynamic spaces, where rescans are recommended. The full pipeline runs locally without cloud upload, supporting privacy-sensitive deployments.